\documentclass{article}

\usepackage{microtype}
\usepackage{graphicx}
\usepackage{subcaption}
\usepackage{booktabs} 

\usepackage{hyperref}

\usepackage[preprint]{icml2027}

\usepackage{amsmath}
\usepackage{amssymb}
\usepackage{mathtools}
\usepackage{amsthm}

\usepackage{colortbl}
\usepackage{tcolorbox}
\usepackage{xurl}
\providecommand{\refcell}[1]{\textcolor{gray}{#1}}
\usepackage[capitalize,noabbrev]{cleveref}

\theoremstyle{plain}

\theoremstyle{definition}

\theoremstyle{remark}

\usepackage[textsize=tiny]{todonotes}

\icmltitlerunning{DuoMI: Expert Reuse in Looped Transformers}

\begin{document}

\twocolumn[
  \icmltitle{DuoMI: Dual Mutual Information Regularization%
  \texorpdfstring{\\}{ }for Expert Reuse in Looped Transformers}


  \icmlsetsymbol{equal}{*}

  \begin{icmlauthorlist}
    \icmlauthor{Xinlai Kang}{huawei}
    \icmlauthor{Zhenhe Wu}{huawei}
    \icmlauthor{Haoyu Wang}{huawei}
    \icmlauthor{Weihua He}{huawei}
    \icmlauthor{Yaping Jin}{huawei}
    \icmlauthor{Jie Hu}{huawei}
    \icmlauthor{Hanting Chen}{huawei}
  \end{icmlauthorlist}

  \icmlaffiliation{huawei}{Huawei Foundation Model Lab}
  
  \icmlcorrespondingauthor{Hanting Chen}{chenhanting@huawei.com}

  \icmlkeywords{Looped Transformers, Mixture of Experts, Cross-Round Expert Routing, Parameter Sharing, Mutual Information Regularization}

  \vskip 0.3in
]



\printAffiliationsAndNotice{}  

\begin{abstract}
We introduce Dual Mutual Information Regularization (DuoMI), a method for regulating expert routing across rounds in looped mixture-of-experts (MoE) Transformers. When two rounds of computation share an expert pool, their routers may favor different expert subsets, limiting the reuse of individual expert parameters across rounds. Even when aggregate expert usage is identical across rounds, we would still prefer each token to retain some experts while switching to others. DuoMI addresses both aspects of routing: it uses round-level marginal mutual information to limit the dependence of aggregate expert usage on the round, and token-level conditional mutual information to penalize average cross-round routing changes that fall outside a target interval. During training, DuoMI computes differentiable penalties from soft routing distributions over all experts. It adds no model parameters and leaves sparse Top-$k$ routing strategy unchanged at inference. On a looped OLMoE backbone in which paired adjacent layers share experts, DuoMI improves the average score across $15$ downstream tasks from $50.08\%$ to $51.82\%$, a gain of $1.74$ percentage points over the looped baseline.
\end{abstract}

\section{Introduction}
\label{sec:introduction}

Improving language model capabilities under limited parameter
and computational budgets is an important objective
in architecture design.
Mixture-of-Experts (MoE) models expand parameter capacity
while controlling per-token computation
by activating only a subset of experts for each token~\citep{shazeer2017sparse,lepikhin2021gshard}.
Looped Transformers offer a complementary approach:
they repeatedly invoke shared parameters at different depths,
enabling multiple rounds of representation transformation
with fewer independent parameters~\citep{dehghani2019universal,geiping2025recurrent}.
Combining sparse expert selection with recurrent parameter sharing
therefore provides a promising way to coordinate model capacity,
computational depth, and parameter storage requirements~\citep{csordas2024moeut}.

This work focuses on looped Transformers with MoE layers.
A central challenge in these architectures is enabling
functional differentiation across rounds that share parameters~\citep{bae2025relaxed,jeddi2026loopformer}.

One architectural approach shares expert FFN parameters
across layers while retaining independently parameterized
attention, router, and normalization modules~\citep{jaggi2026tying}.

Architectural mechanisms for differentiation can benefit
from training signals that encourage their intended
functional roles to emerge.
In looped MoE architectures, more precise control over
how shared experts are actually reused may help realize
the potential of these designs.
Although the same expert pool is invoked in both rounds,
the routers may develop different expert selection preferences~\citep{lee2026sparse}.
At the round level, the two rounds may favor different
expert subsets, effectively partitioning the shared pool
according to invocation round.
At the token level, even identical aggregate usage distributions
can coexist with unchanged, partially overlapping,
or entirely disjoint expert assignments for individual tokens.

\begin{figure*}[t]
    \centering
    \includegraphics[width=\textwidth]{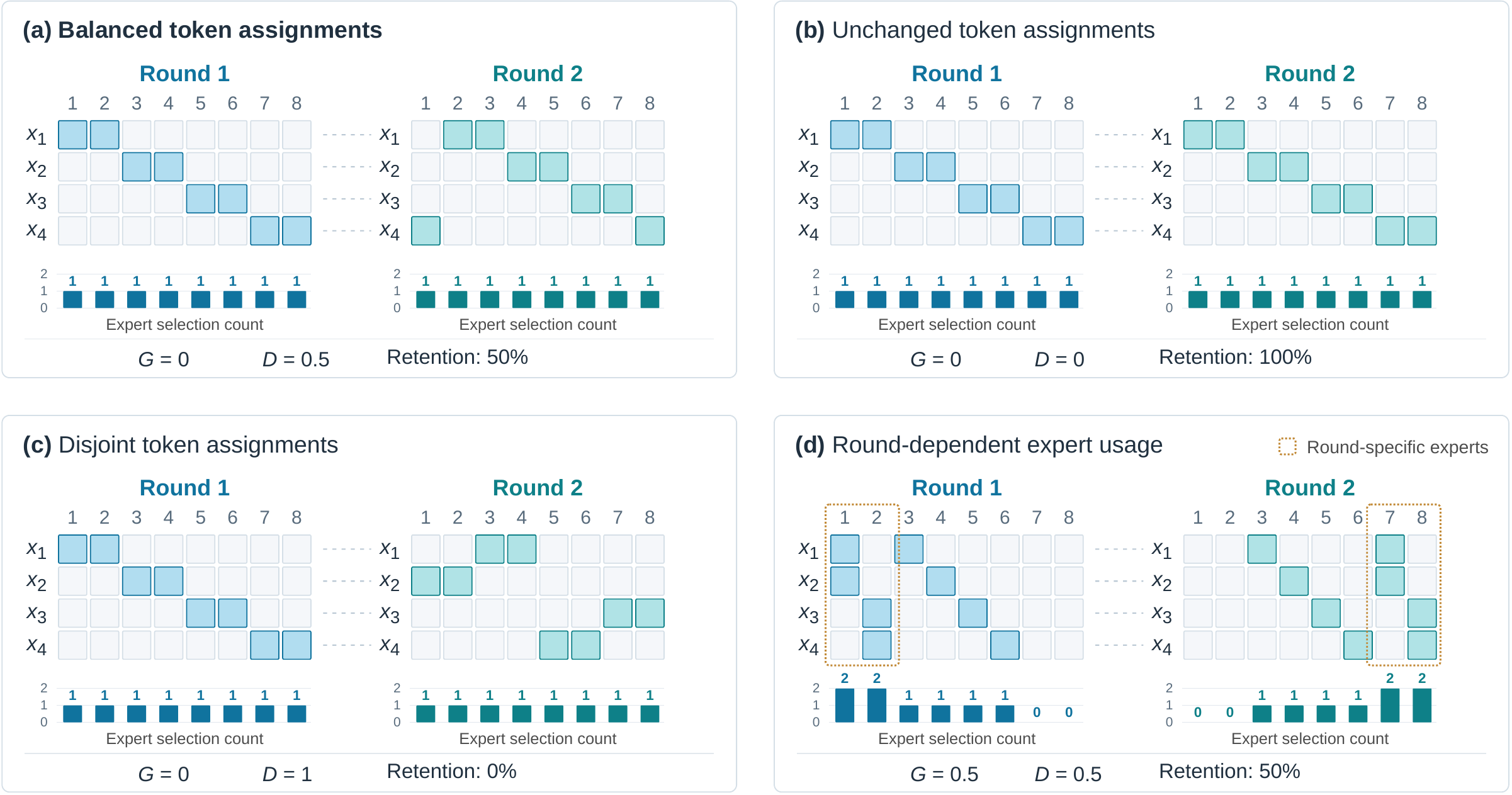}
    \caption{
        Cross-round expert selection patterns.
        Rows denote tokens and columns denote experts;
        colored cells indicate selected experts.
        The bar charts show expert selection counts
        in each round.
        Both $G$ and $D$ are computed using uniform
        hard Top-$k$ distributions.
        (a) The aggregate distributions are identical,
        and each token retains half of its experts:
        $G=0,D=0.5$.
        (b) Each token retains its entire expert set:
        $G=0,D=0$.
        (c) Each token selects disjoint expert sets
        across rounds:
        $G=0,D=1$.
        (d) Each token again retains half of its experts,
        but the rounds favor different expert subsets:
        $G=0.5,D=0.5$.
        These patterns illustrate aggregate expert reuse
        and token-level routing change as distinct
        statistical dimensions; they do not specify
        the soft regularization thresholds used in training.
    }
    \label{fig:gd_intro}
\end{figure*}

Figure~\ref{fig:gd_intro} illustrates the distinction
between these two statistical levels.
For clarity, we consider a simplified setting with $8$ experts
and $2$ experts selected per token.
Here, $G$ measures the difference between the aggregate
expert usage distributions of the two rounds,
and $D$ measures the average fraction of selected experts
replaced across rounds for the same token.
In Figures~\ref{fig:gd_intro}(a) and~\ref{fig:gd_intro}(d),
every token retains half of its selected experts,
giving $D=0.5$ in both cases.
However, the aggregate distributions are identical
in the former, yielding $G=0$,
whereas the latter includes experts used exclusively
in one round, yielding $G=0.5$.
Thus, controlling token-level expert retention alone
may still allow round-dependent usage preferences
within a shared pool.
Constraining differences in aggregate usage
encourages the same expert parameters to be used
across rounds.

Conversely, Figures~\ref{fig:gd_intro}(a)--(c)
have identical aggregate expert usage distributions
across rounds, with perfectly balanced selection counts
within each round.
Nevertheless, their token-level routing patterns range
from complete retention to complete replacement.
Aggregate distribution alignment therefore does not determine
how individual tokens use shared experts across rounds.
Unchanged selections preserve a stable expert combination
for each token, but do not adapt the membership of that
combination across layers.
Complete replacement changes the combination entirely,
but retains no experts for the same token across rounds.
These observations motivate treating routing change
as a quantity whose degree should be controlled
through both a lower and an upper bound.
Importantly, an expert receives different hidden states
in the two rounds, so selecting it again can still
perform a different computation.
We therefore do not equate expert overlap
with computational redundancy.
Instead, we empirically examine whether regulating
the degree of routing change improves learning
in shared-expert architectures.

This objective differs from the uniformity of expert usage
encouraged by load balancing~\citep{fedus2022switch}.
Figures~\ref{fig:gd_intro}(a)--(c)
have the same uniform hard expert usage distributions,
yet exhibit different token-level routing behavior.
Conversely, both rounds can favor the same small subset
of experts and still have identical aggregate distributions.
Cross-round distribution alignment is therefore
not equivalent to uniform usage across all experts.
Our goal is to retain the original load-balancing loss
while additionally modeling cross-round relationships
in aggregate expert usage and in the expert assignments
of individual tokens.

Regulating actual expert reuse in looped Transformers
thus requires considering both aggregate usage distributions
and token-level routing correspondence.
This leads to our central question:

\begin{center}
\begin{tcolorbox}[colframe=black!50!white, 
    colback=gray!5!white, 
    boxrule=0.5mm, 
    arc=5pt,width=8cm]
\centering
\textit{How can we promote aggregate expert reuse across rounds while preserving a moderate degree of expert selection change for individual tokens?}
\end{tcolorbox}
\end{center}

To address this question, we propose
\emph{Dual Mutual Information Regularization} (DuoMI).
We represent the round, expert, and token position
with its causal context by random variables
$R$, $E$, and $X$, respectively,
and characterize cross-round routing using
two mutual information quantities.
Round-level marginal mutual information,
$G=I(R;E)$,
measures the dependence of aggregate expert usage
on the invocation round;
an upper-bound constraint discourages round-specific
partitioning of the shared pool.
Token-level conditional mutual information,
$D=I(R;E\mid X)$,
measures routing change conditioned on
the token-context pair;
a two-sided constraint limits both insufficient
and excessive changes.
Under the empirical distribution we define,
these quantities satisfy $0\leq G\leq D\leq1$,
while a small $G$ can coexist with a large $D$.
This formulation jointly characterizes aggregate reuse
and token-level change within a unified
information-theoretic framework,
without collapsing them into a single expert overlap metric.

Statistics based on hard Top-$k$ sets have zero gradients
with respect to router logits almost everywhere.
DuoMI therefore constructs differentiable mutual information
surrogates from soft routing distributions over all experts
and implements the upper-bound and interval regularization
using squared hinge penalties.
The additional regularizer contributes only to
the training objective.
It introduces no model parameters and leaves
the original sparse Top-$k$ routing,
expert output weighting, and inference computation unchanged.
By providing cross-round training signals
to the existing routers,
DuoMI encourages shared use of expert parameters
while preserving flexibility to adapt expert selection
to the hidden states at different layers.

We conduct main comparisons, ablations,
and threshold selection experiments
on a shared-expert architecture built from OLMoE~\citep{muennighoff2025olmoe}.
We also analyze routing behavior using both
soft mutual information statistics
and expert retention measured from actual Top-$k$ selections.
Under our main experimental setting,
DuoMI improves the average score across $15$ downstream tasks
from $50.08\%$ for the OLMoE-Loop baseline
to $51.82\%$, a gain of $1.74$ percentage points.
We further compare alternative routing adaptations
on the same backbone and examine individual components
by removing mutual information terms or interval bounds.
Our empirical conclusions focus on this shared-expert
architecture and assess the effectiveness
of cross-round routing regularization within this setting.

Our main contributions are as follows:
\begin{itemize}
    \item
    We characterize cross-round expert reuse
    at two levels: aggregate expert usage
    and token-level routing change.
    Marginal and conditional mutual information
    provide a unified statistical description
    that clarifies their relationship and distinction.

    \item
    We propose DuoMI to jointly regulate
    cross-round differences in aggregate expert usage
    and the range of token-level routing change.
    Differentiable surrogates enable training
    without introducing model parameters
    or modifying the inference routing procedure.

    \item
    We conduct comparative, ablation,
    and threshold experiments on an OLMoE
    shared-expert backbone to evaluate performance gains
    within this setting.
    We further combine soft statistics
    with hard routing metrics to analyze
    how the regularizer affects expert selection.
\end{itemize}


\section{Related Work}
\label{sec:related_work}

\paragraph{Recurrent parameter sharing and differentiation across rounds.}
Sharing parameters across layers allows Transformers to increase effective
computational depth while controlling the number of unique parameters.
Universal Transformers recurrently apply shared transformations and incorporate
iteration signals and adaptive halting~\citep{dehghani2019universal},
while ALBERT demonstrates the parameter efficiency of cross-layer sharing
in language representation learning~\citep{lan2020albert}.
To preserve flexibility across depths, Relaxed Recursive Transformers introduce
layer-specific low-rank adaptations~\citep{bae2025relaxed},
and LoopFormer combines conditioning on time and step size with
shortcut-consistency training to coordinate recurrent trajectories
of different lengths~\citep{jeddi2026loopformer}.
Mixture-of-Recursions further assigns different recursion depths to individual
tokens, enabling adaptive allocation of computation~\citep{bae2025recursions}.
These approaches improve shared computation through parameter adaptation,
round-dependent conditioning, or depth allocation.
DuoMI focuses on the selection of shared experts, using training regularization
to control how they are used across two rounds.

\paragraph{Looped MoE and expert parameter sharing.}
Sparse MoE models coordinate parameter capacity and computational cost
by selecting a subset of experts for each input, with Switch Transformers
and OLMoE providing representative training and model designs~\citep{fedus2022switch,muennighoff2025olmoe}.
MoEUT combines expert mechanisms with layer sharing in Universal
Transformers~\citep{csordas2024moeut}.
LoopMoE integrates sparse routing with iterative computation through IterAdaLN,
which conditions on both the iteration and token state, together with
capacity balancing~\citep{chen2026loopmoe}.
Chain-of-Experts uses a separate router at each step to select experts sequentially within a single layer~\citep{wang2025chainexperts}.
Regarding expert reuse, Tying the Loop shares expert FFNs across adjacent layers
while retaining separate attention, router, and normalization parameters,
making it directly relevant to the sharing scheme used in this
work~\citep{jaggi2026tying}.
MoRE allows adjacent layers to select from a larger shared expert pool,
using independent routers and depth embeddings to distinguish
layers~\citep{qiu2026more}.
Foil studies the configuration of expert pool width, loop depth,
and attention sharing; its design shares experts and routers while retaining
separate attention parameters for each pass~\citep{wang2026foil}.
MoUE expands the effective space of expert combinations through reusable
cross-layer expert pools, structured connectivity, load balancing that accounts
for differences in expert exposure, and stateful routing~\citep{chen2026moue}.
Meta Expert studies architectures that combine a pool shared across layers with experts private to individual layers~\citep{hua2026metaexpert}.
These studies establish several architectural approaches to sharing experts.
We adopt a shared-expert backbone and focus on cross-round routing
regularization within a fixed architecture.

\paragraph{Cross-layer routing coordination and cross-round behavior.}
RMoE uses a GRU to propagate states between routers at different layers
and analyzes routing dependence through cross-layer mutual
information~\citep{qiu2025rmoe}.
In that work, mutual information serves as a diagnostic rather than an explicit
training regularizer.
PathMoE shares router parameters across consecutive layers to constrain
the effective expert-path space~\citep{gu2026pathmoe}.
HERO-MoE incorporates earlier layers' detached routing distributions into the current router logits~\citep{qiu2026hero}.
Beyond architectural modifications, \citet{hu2026synergistic} combine a loss
that penalizes similarity between expert activations with a cross-layer routing
coupling loss, promoting intra-layer specialization and coordinated paths
across layers, respectively.
SARA averages soft routing probabilities within each sequence and uses
Jensen--Shannon divergence to align the aggregate routing distribution
of low-resource language inputs with high-resource language
anchors~\citep{dong2026sara}.
Its formulation of averaging before comparing distributions is closely related
to our $\widetilde G$, but aligns inputs across languages rather than repeated
invocations of the same shared expert pool.
For looped architectures, \citet{lee2026sparse} analyze changes in expert
assignments for the same token across rounds by distinguishing fully
overlapping, partially overlapping, and disjoint expert sets.
LoopMoE also uses the Jaccard similarity of selected expert sets to examine
stability and change at different stages of recurrent
computation~\citep{chen2026loopmoe}.
LOOM uses loop-specific routers with shared experts and reports lower cosine similarity between aggregate expert loads across loops~\citep{he2026loom}.
These analyses provide evidence about routing behavior in looped MoE models,
but observing cross-round differences does not establish their optimal degree.
DuoMI separately models aggregate distribution differences and token-level
changes, imposing both a lower and an upper bound on the latter.

\paragraph{Mutual information for expert selection.}
Mutual information has been used to construct MoE training objectives.
Mod-Squad learns task-dependent expert combinations through mutual information
between tasks and experts~\citep{chen2023modsquad}.
ModuleFormer and DS-MoE use input--expert mutual information objectives,
combining high marginal expert entropy with low conditional routing entropy
to coordinate aggregate usage and concentrated selection for individual
inputs~\citep{shen2023moduleformer,pan2024dsmoe}.
Under balanced marginal expert usage, Hi-MoE relates its within-group regularizer to collision mutual information between tokens and experts~\citep{molodtsov2026himoe}.
Conditional mutual information has also been applied to MoE training:
M4oE uses it to learn modality--task dependencies in multimodal medical
applications~\citep{wu2025m4oe}.
These methods differ from DuoMI in their random variables and target
dependence structures.
Specialization driven by inputs, tasks, or modalities does not directly specify
how routing should change when the same token repeatedly invokes shared experts.

Within a shared pool whose expert identities are aligned across rounds,
we use $G=I(R;E)$ and $D=I(R;E\mid X)$ to characterize aggregate dependence
of expert usage on the round and routing change conditioned on the
token-context pair, respectively.
DuoMI penalizes excessive $G$ and deviations of $D$ from a target interval,
using differentiable surrogates while retaining the original task
and load-balancing objectives.
Our contribution lies in jointly regulating these two statistical levels,
rather than in expert sharing, mutual information regularization,
or routing distribution alignment alone.

\section{Method}
\label{sec:method}
We propose \emph{Dual Mutual Information Regularization} (DuoMI)
to regulate cross-round expert selection in looped Transformers.
DuoMI jointly controls round-level expert usage distributions
and token-level routing changes,
shaping how shared experts are activated across rounds.
The additional regularization terms are computed from router outputs
and incorporated into the training objective,
while leaving the original Top-$k$ routing mechanism
and expert output weighting unchanged.

\begin{figure}[ht]
    \centering
    \includegraphics[width=0.8\linewidth]{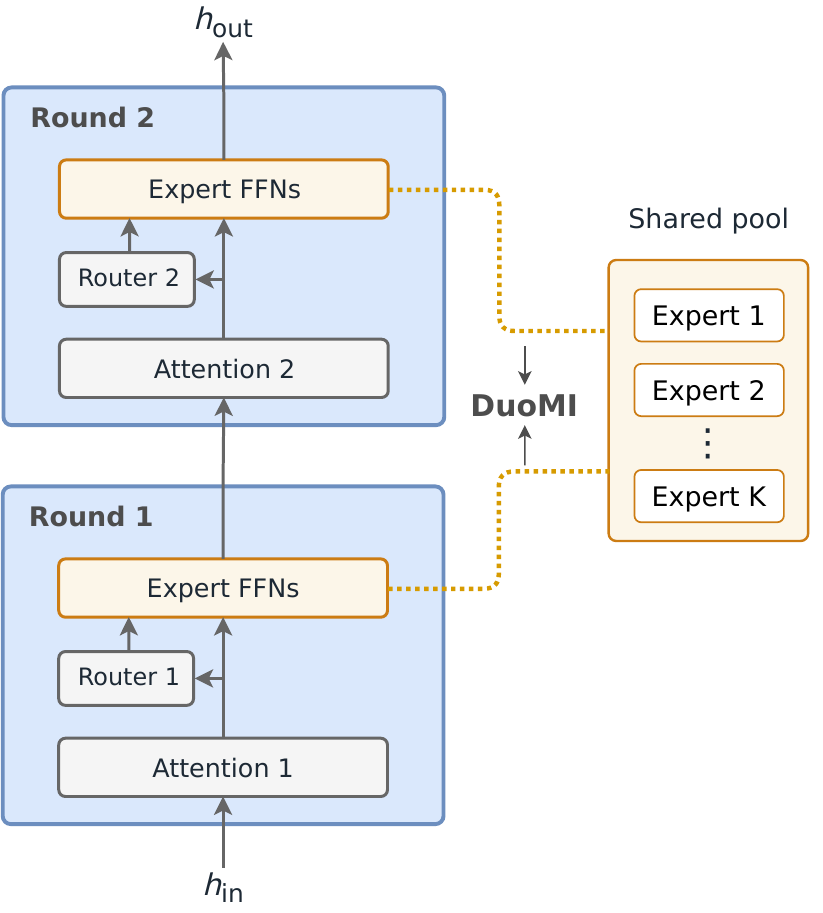}
    \caption{
        Expert sharing in the looped Transformer backbone.
        The two rounds share the same pool of expert FFNs,
        while attention, router, and normalization parameters
        remain independent across rounds.
    }
    \label{fig:architecture}
\end{figure}

As illustrated in Figure~\ref{fig:architecture},
we consider a Transformer with $M$ shared expert pools.
Each pool contains $K$ experts and is invoked at two distinct layers.
The two invocations share expert FFN parameters,
while the attention, router, and normalization modules
remain independently parameterized~\citep{jaggi2026tying}.
We refer to these two invocations of the same expert pool
as two \emph{rounds}, indexed by $r\in\{1,2\}$.
Here, recurrent reuse is restricted to expert parameters
and does not require parameter sharing across entire Transformer blocks.
In our OLMoE configuration, there are $M=6$ shared expert pools,
each containing $K=64$ experts,
with $k=8$ experts selected per token at each invocation.

Sharing expert parameters does not, by itself,
specify how the two rounds should use the same pool.
Over a given set of token positions,
the two rounds may favor different expert subsets in aggregate.
Even when their aggregate usage distributions are similar,
the expert assignments for individual tokens may differ across rounds.
Moreover, the same expert receives different hidden states
in the two rounds,
so selecting it again does not necessarily repeat the same computation.
These observations motivate jointly modeling and regulating
two complementary aspects of cross-round routing:
round-level expert usage and token-level routing changes.

\subsection{Round-Level Marginal Mutual Information for Expert Reuse}
\label{sec:round_level_mi}

For a statistical group $b$, let
$\mathcal{B}_b=\{x_i\}_{i=1}^{N_b}$
denote its $N_b>0$ valid token positions together with their
causal contexts.
Statistics from the two rounds are aligned to the same token
positions in the same input sequences,
although the corresponding hidden states may differ across layers.
Let $S_{j,r}(x)$ denote the Top-$k$ expert set selected for
token-context pair $x$ by expert pool $j$ in round $r$.
To characterize expert selection independently of output weighting,
we define a uniform distribution over the selected experts:
\begin{equation}
q_{j,r}(e\mid x)
=
\frac{\mathbf{1}\{e\in S_{j,r}(x)\}}{k},
\qquad e\in\{1,\ldots,K\}.
\label{eq:duomi_hard_distribution}
\end{equation}
This distribution is used solely to define routing statistics
and does not replace the expert output weights used by the model.

For a fixed statistical group $b$ and expert pool $j$,
let $X$ be uniformly distributed over the valid token positions
in the group.
Let the round variable $R$ be uniformly distributed over
$\{1,2\}$ and independent of $X$.
Conditioned on $X=x$ and $R=r$,
the expert variable $E$ is sampled from
$q_{j,r}(\cdot\mid x)$.
The resulting empirical joint distribution is
\begin{equation}
P_{b,j}(X=x_i,R=r,E=e)
=
\frac{1}{2N_b}q_{j,r}(e\mid x_i).
\label{eq:duomi_joint_distribution}
\end{equation}
The average expert usage distribution in round $r$ is therefore
\begin{equation}
\bar q_{b,j,r}(e)
=
\frac{1}{N_b}
\sum_{i=1}^{N_b}q_{j,r}(e\mid x_i).
\label{eq:duomi_hard_marginal}
\end{equation}

We use the marginal mutual information between the round
and expert variables to measure how strongly aggregate expert
usage depends on the invocation round:
\begin{equation}
G_{b,j}
=
I(R;E)
=
\operatorname{JSD}_2
\left(
\bar q_{b,j,1},
\bar q_{b,j,2}
\right).
\label{eq:duomi_global_mi}
\end{equation}
Here and throughout, mutual information is evaluated under
the corresponding distribution $P_{b,j}$.
For two probability distributions $a$ and $c$,
let $m=(a+c)/2$.
Their Jensen--Shannon divergence~\citep{lin1991divergence} is
\begin{equation}
\operatorname{JSD}_2(a,c)
=
\frac{1}{2}\operatorname{KL}_2(a\Vert m)
+
\frac{1}{2}\operatorname{KL}_2(c\Vert m).
\label{eq:duomi_jsd}
\end{equation}
All logarithms are base $2$,
and contributions from zero-probability terms are defined
as zero by continuity.

A small $G_{b,j}$ indicates similar aggregate expert usage
across the two rounds,
whereas a large value indicates round-dependent preferences
within the shared pool.
Imposing an upper bound on $G_{b,j}$ therefore discourages
the pool from partitioning into round-specific expert subsets
and promotes cross-round reuse of the same expert parameters.
Intuitively, if the two rounds select entirely disjoint
expert subsets, the shared pool effectively functions
as two separate, round-specific subpools.
Although parameter sharing still reduces the model's
parameter count, the selected experts are not actually
reused across rounds.
Our constraint addresses this gap between parameter sharing
and realized cross-round expert reuse.

A small $G_{b,j}$ does not, however, require uniform expert usage.
For example, $G_{b,j}=0$ whenever the two rounds have identical
aggregate usage distributions,
even if both concentrate on the same small subset of experts.
Thus, this constraint regulates cross-round alignment
rather than uniformity across experts
and does not replace the original load-balancing loss.

\subsection{Token-Level Conditional Mutual Information for Routing Change}
\label{sec:token_level_mi}

Comparing aggregate expert usage across rounds does not
directly constrain cross-round routing for individual tokens.
We therefore consider the conditional mutual information
between the round and expert variables,
conditioned on the token-context pair:
\begin{equation}
\begin{aligned}
D_{b,j}
&=
I(R;E\mid X) \\
&=
\frac{1}{N_b}
\sum_{i=1}^{N_b}
\operatorname{JSD}_2
\left(
q_{j,1}(\cdot\mid x_i),
q_{j,2}(\cdot\mid x_i)
\right).
\end{aligned}
\label{eq:duomi_conditional_mi}
\end{equation}
Whereas $G_{b,j}$ compares distributions after averaging
over tokens, $D_{b,j}$ compares the two routing distributions
for each token before averaging over the valid token
positions in the group.

Under the uniform Top-$k$ distributions defined
in Eq.~\eqref{eq:duomi_hard_distribution},
$D_{b,j}$ admits a direct interpretation as
the average fraction of selected experts replaced across rounds:
\begin{equation}
D_{b,j}
=
1-
\frac{1}{N_b}
\sum_{i=1}^{N_b}
\frac{
\left|
S_{j,1}(x_i)\cap S_{j,2}(x_i)
\right|
}{k}.
\label{eq:duomi_replacement}
\end{equation}
Experts selected in both rounds contribute zero to the JSD;
only experts selected in exactly one round contribute
to the divergence.
Consequently, $D_{b,j}$ equals the fraction of experts
replaced between the two rounds, averaged over tokens.
It is zero when the selected sets are identical
across rounds for every token,
and one when these sets are disjoint for every token.

We impose a two-sided constraint on this degree of routing change.
The lower bound discourages overly similar expert selection
across rounds, while the upper bound limits excessive changes
to promote routing stability.
This constraint applies to the group-level average
and does not require every token to replace the same number
or fraction of experts.

\subsection{Dual Mutual Information Regularization}
\label{sec:dual_mi_regularization}

Under the empirical joint distribution defined above,
the two mutual information quantities satisfy
\begin{equation}
0\leq G_{b,j}\leq D_{b,j}\leq 1.
\label{eq:duomi_information_bounds}
\end{equation}
The upper bound follows from the entropy of the binary
round variable, $H(R)=1$.
Since $R$ and $X$ are independent,
the chain rule for mutual information~\citep[Sec.~2.5]{cover2006elements} further gives
\begin{equation}
D_{b,j}-G_{b,j}
=
I(R;X\mid E)
\geq 0.
\label{eq:duomi_information_gap}
\end{equation}

Despite this relationship, a small $G_{b,j}$
does not imply a small $D_{b,j}$.
For example, the two rounds can have identical aggregate
expert usage distributions while replacing half of
the selected experts for every token.
In this case, $G_{b,j}=0$ and $D_{b,j}=1/2$.
Thus, aggregate expert reuse can coexist
with token-level routing changes.
Conversely, constraining $D_{b,j}$ to an interval
does not generally ensure that $G_{b,j}$ reaches
the desired low level.
Controlling token-level changes alone may still allow
the two rounds to favor different expert subsets in aggregate.

Motivated by this complementarity,
we combine an upper-bound constraint on round-level
marginal mutual information with a two-sided constraint
on token-level conditional mutual information.
The former discourages round-specific partitioning
of aggregate expert usage,
promoting reuse of the same expert parameters across rounds.
The latter controls the average degree of cross-round
routing change for individual tokens.
Together, these constraints allow the model to reuse
shared experts while adapting token-level expert selection
to the hidden states at different layers.

\subsection{Differentiable Surrogates and Training Objective}
\label{sec:differentiable_training}

\paragraph{Differentiable routing distributions.}

The hard routing statistics defined above depend on
membership in the Top-$k$ expert sets,
rather than the continuous weights assigned to selected experts.
For a given token-context pair $x$,
the selected set $S_{j,r}(x)$ remains unchanged
as long as changes in the logits do not alter
which experts fall within the Top-$k$.
Consequently, $G_{b,j}$, $D_{b,j}$,
and the corresponding regularization penalties
are piecewise constant with respect to the router logits,
with zero gradients almost everywhere~\citep{xie2020differentiable}.
Direct backpropagation through these hard statistics
therefore provides no useful gradient signal for optimization.

To enable gradient-based optimization,
we construct differentiable surrogates using
the router's relative preferences over all experts.
Specifically, given the router logits
$z_{j,r}(x)\in\mathbb{R}^{K}$,
we define
\begin{equation}
p_{j,r}(e\mid x)
=
\frac{
\exp\!\left(z_{j,r,e}(x)/\tau\right)
}{
\sum_{e'=1}^{K}
\exp\!\left(z_{j,r,e'}(x)/\tau\right)
},
\qquad \tau>0.
\label{eq:duomi_soft_distribution}
\end{equation}
The temperature $\tau$ controls the smoothness
of this auxiliary distribution.
Unlike $q_{j,r}$, which depends on discrete set membership,
$p_{j,r}$ varies smoothly with the logits.
Its partial derivative is
\begin{equation}
\frac{
\partial p_{j,r}(e\mid x)
}{
\partial z_{j,r,e'}(x)
}
=
\frac{1}{\tau}
p_{j,r}(e\mid x)
\left(
\mathbf{1}\{e=e'\}
-
p_{j,r}(e'\mid x)
\right).
\label{eq:duomi_softmax_gradient}
\end{equation}
Thus, even when a logit update leaves the hard Top-$k$
set unchanged, the auxiliary distribution can reflect
changes in relative expert preferences,
allowing the resulting regularizer to provide gradient signals.
Because the softmax spans all $K$ experts,
these gradients can also reach the logits of experts
that are not currently selected.

For statistical group $b$,
we define the average soft routing distribution
in each round as
\begin{equation}
\bar p_{b,j,r}(e)
=
\frac{1}{N_b}
\sum_{i=1}^{N_b}
p_{j,r}(e\mid x_i).
\label{eq:duomi_soft_marginal}
\end{equation}
Replacing the hard distributions with their soft counterparts
yields two differentiable statistics:
\begin{equation}
\begin{aligned}
\widetilde G_{b,j}
&=
\operatorname{JSD}_2
\left(
\bar p_{b,j,1},
\bar p_{b,j,2}
\right),\\
\widetilde D_{b,j}
&=
\frac{1}{N_b}
\sum_{i=1}^{N_b}
\operatorname{JSD}_2
\left(
p_{j,1}(\cdot\mid x_i),
p_{j,2}(\cdot\mid x_i)
\right).
\end{aligned}
\label{eq:duomi_soft_statistics}
\end{equation}
Here, $\widetilde G_{b,j}$ first averages routing probabilities
over tokens and then compares the two round-level distributions.
In contrast, $\widetilde D_{b,j}$ first compares
the two routing distributions for each token
and then averages over valid token positions.
These statistics preserve the distinction between
aggregate expert usage and token-level routing changes.
Under the auxiliary probabilistic model in which
the expert variable is sampled from $p_{j,r}$,
$\widetilde G_{b,j}$ and $\widetilde D_{b,j}$
remain the marginal and conditional mutual information,
respectively, between the round and expert variables.

These soft statistics serve as surrogates for
the hard routing objectives;
they are neither exact replacements for the hard metrics
nor exact estimators of their gradients.
The hard distributions encode only which experts are selected,
whereas the soft distributions retain information
about the relative scores of all experts.
Changes in the logits can therefore alter
$\widetilde G_{b,j}$ and $\widetilde D_{b,j}$
even when the hard selection sets remain unchanged.
During training, we also monitor hard metrics computed
from the actual Top-$k$ selections
to assess how soft regularization affects routing behavior.

\paragraph{Dual mutual information regularization.}

Let $[a]_+=\max(a,0)$ and
$\lambda_G,\lambda_L,\lambda_U\geq0$.
For each statistical group and expert pool,
we define the squared hinge penalty
\begin{equation}
\begin{aligned}
\Phi_{b,j}
={}&
\lambda_G
\left[
\widetilde G_{b,j}
-
\widetilde g_{\max}
\right]_+^2\\
&+
\lambda_L
\left[
\widetilde d_{\min}
-
\widetilde D_{b,j}
\right]_+^2\\
&+
\lambda_U
\left[
\widetilde D_{b,j}
-
\widetilde d_{\max}
\right]_+^2,
\end{aligned}
\label{eq:duomi_penalty}
\end{equation}
where
$0\leq\widetilde g_{\max}\leq1$
and
$0\leq\widetilde d_{\min}
\leq\widetilde d_{\max}\leq1$.
The first term penalizes excessive differences
in aggregate expert usage across rounds.
The second and third terms penalize insufficient
and excessive token-level routing changes, respectively.
When $\widetilde D_{b,j}$ lies within the target interval,
both associated penalties vanish,
so the regularizer does not continually push this statistic
toward either endpoint.
These finite-weight soft penalties express optimization
preferences rather than guaranteeing strict threshold
satisfaction for every statistical group.

\paragraph{Statistical grouping and training objective.}

Each mutual information estimate uses all valid token positions
from a fixed group of $8$ training sequences.
Statistics are computed separately for each expert pool,
with exactly the same valid token positions used
in both rounds.
We fix the softmax temperature to $\tau=1.0$.

Let $\mathcal{G}_t$ denote the set of statistical groups
included in optimizer update $t$,
and let $B_t=|\mathcal{G}_t|$.
Penalties are computed separately for each group and pool
and then averaged.
The complete training objective is
\begin{equation}
\mathcal{L}_t
=
\mathcal{L}_{\mathrm{base},t}
+
\frac{w_t}{B_t M}
\sum_{b\in\mathcal{G}_t}
\sum_{j=1}^{M}
\Phi_{b,j},
\label{eq:duomi_total_loss}
\end{equation}
where
\begin{equation}
\mathcal{L}_{\mathrm{base},t}
=
\mathcal{L}_{\mathrm{LM},t}
+
\alpha_{\mathrm{bal}}
\mathcal{L}_{\mathrm{bal},t}
+
\alpha_z
\mathcal{L}_{z,t}.
\label{eq:duomi_base_loss}
\end{equation}
The base objective comprises the language-modeling loss,
load-balancing loss, and router z-loss,
with their coefficients unchanged from the baseline.
The factor $w_t$ controls the schedule
of the additional regularization:
we set $w_t=1$ when no schedule is used,
or gradually increase it to $1$ during regularization warmup.
This schedule is configured independently
of learning-rate warmup.
Algorithm~\ref{alg:duomi_training} summarizes
the computation of DuoMI during training.

\floatstyle{ruled}
\restylefloat{algorithm}
\renewcommand{\algorithmicrequire}{\textbf{Input:}}
\renewcommand{\algorithmicensure}{\textbf{Output:}}

\begin{algorithm}[ht]
\caption{Dual Mutual Information Regularization}
\label{alg:duomi_training}
\small
\begin{algorithmic}[1]

\REQUIRE
Parameters $\theta$;
microbatches $\{\mathcal B_t^{(a)}\}_{a=1}^{A_t}$;
group size $s=8$;
$M$, $\tau$, $w_t$, and penalty function $\Phi$.

\STATE Zero gradients.

\FOR{$a=1,\ldots,A_t$}

    \STATE
    $\mathcal L_{\mathrm{base}}^{(a)},
    \{z_{j,r}\}
    \gets \operatorname{Forward}_{\theta}
    (\mathcal B_t^{(a)})$.

    \STATE
    $\mathcal G_a
    \gets \operatorname{GroupSequences}
    (\mathcal B_t^{(a)},s)$.

    \STATE $R_a \gets 0$.

    \FOR{each group $b\in\mathcal G_a$}

        \STATE
        $\{x_i\}_{i=1}^{N_b}
        \gets \operatorname{ValidTokens}(b)$.

        \FOR{$j=1,\ldots,M$}

            \STATE
            $p_{i,r}
            \gets
            \operatorname{softmax}
            (z_{j,r}(x_i)/\tau),
            \quad \forall i,\ r\in\{1,2\}$.

            \STATE
            $\bar p_r
            \gets N_b^{-1}\sum_i p_{i,r},
            \quad r\in\{1,2\}$.

            \STATE
            $\widetilde G_{b,j}
            \gets
            \operatorname{JSD}_2(\bar p_1,\bar p_2)$.

            \STATE
            $\widetilde D_{b,j}
            \gets
            N_b^{-1}\sum_i
            \operatorname{JSD}_2(p_{i,1},p_{i,2})$.

            \STATE
            $R_a
            \gets
            R_a+\Phi(\widetilde G_{b,j},
                     \widetilde D_{b,j})$.

        \ENDFOR

    \ENDFOR

    \STATE
    $\mathcal L_{\mathrm{MI}}^{(a)}
    \gets R_a/(M|\mathcal G_a|)$.

    \STATE
    $\ell_a
    \gets
    \bigl(
    \mathcal L_{\mathrm{base}}^{(a)}
    +w_t\mathcal L_{\mathrm{MI}}^{(a)}
    \bigr)/A_t$.

    \STATE Backpropagate $\ell_a$ and accumulate gradients.

\ENDFOR

\STATE Update $\theta$ using the accumulated gradients.

\end{algorithmic}
\end{algorithm}

\section{Experiments}
\label{sec:experiments}

We evaluate DuoMI on a looped OLMoE backbone with shared expert pools.
We first compare DuoMI with alternative routing methods on downstream tasks.
We then assess the contribution of each component by removing the round-level
constraint, the token-level constraint, or either bound of the token-level
target band in turn. Finally, we examine routing trajectories during training
and vary the regularization thresholds to study how DuoMI changes actual
Top-$k$ expert selections and how these changes relate to downstream performance.

\subsection{Experimental Setup}
\label{sec:experimental_setup}

\paragraph{Model and training setup.}
Our main experiments use an OLMoE~\citep{muennighoff2025olmoe} backbone in which adjacent layers
share experts. The model has $16$ layers. The first two and last two
layers each retain their own experts, while the middle $12$ layers are
arranged into six adjacent pairs, yielding $M=6$ shared expert pools.
Each pool contains $K=64$ experts, of which $k=8$ are selected per token
at each invocation. Counting shared weights only once, the looped model
has approximately $4.5$ billion trainable parameters, with approximately
$1.3$ billion active per token. In both DuoMI and its direct baseline,
the two layers in each pair share only the expert FFN parameters;
the attention, router, and normalization parameters are separate for
each layer. DuoMI therefore has the same parameter count and performs
the same inference computations as the OLMoE-Loop baseline.

We train all looped models from scratch on the same English corpus,
using the official \texttt{sample-100BT} subset of FineWeb-Edu v1.4.0~\citep{penedo2024fineweb}.
All models use the same training budget and downstream evaluation
protocol. Each model is trained for $50{,}000$ optimizer steps with a
sequence length of $4{,}096$ and a global batch size of $512$ sequences,
processing just over $104$ billion token positions in total.

For DuoMI, each mutual information estimate uses a group of eight
training sequences. Within each shared expert pool, we align the valid
token positions across the two rounds. We compute the regularization
penalty separately for each group and pool, then average the penalties
over both. By default, we set $\tau=1$, $\tilde g_{\max}=0.02$, and
$[\tilde d_{\min},\tilde d_{\max}]=[0.05,0.30]$.
The three squared hinge penalties have equal weights:
$\lambda_G=\lambda_L=\lambda_U=0.1$.

\paragraph{Comparison methods.}
Our primary comparison is the \emph{OLMoE-Loop baseline}, which uses
the same shared experts without cross-round routing regularization.
We also adapt three routing methods to the same backbone.
\emph{ModuleFormer-style MI}~\citep{shen2023moduleformer} optimizes
mutual information between inputs and experts separately at each
invocation. The \emph{Path-style shared router}~\citep{gu2026pathmoe}
shares router parameters across the two invocations of an expert pool
but selects experts separately from the hidden states at each round.
The \emph{RMoE-style recurrent router}~\citep{qiu2025rmoe} passes a
routing state from the first invocation to the second, allowing the
second routing decision to use information from the first.
These adaptations compare routing mechanisms within the looped OLMoE
architecture; they do not reproduce the full architectures of the
original papers. We also report the \emph{OLMoE baseline}, which does
not share experts across layers, as an additional reference and exclude
it from the ranking of looped models.

\paragraph{Evaluation metrics.}
We evaluate all models on the $15$ downstream tasks listed in
Table~\ref{tab:duomi_main}, using the metric associated with each task.
For example, we report exact match for BBH~\citep{suzgun2023bbh} and TriviaQA~\citep{joshi2017triviaqa}, normalized
accuracy for ARC-C and ARC-E~\citep{clark2018arc}, and accuracy for BoolQ~\citep{clark2019boolq} and MMLU~\citep{hendrycks2021mmlu}.
All scores are reported as percentages. The \emph{Avg.} column gives
the unweighted arithmetic mean across the $15$ tasks. All score
differences below are expressed in percentage points.

\subsection{Main Results}
\label{sec:main_results}

\begin{table*}[t]
\centering
\caption{
Benchmark comparison on 15 downstream tasks.
All scores are percentages; higher is better.
Bold indicates the best score in each column among the looped models.
OLMoE baseline is shown in gray as an additional reference;
its scores are bolded only when they attain the overall column maximum.
}
\label{tab:duomi_main}

\begingroup
\setlength{\tabcolsep}{2pt}
\renewcommand{\arraystretch}{1.15}
\small

\resizebox{\textwidth}{!}{%
\begin{tabular}{l*{16}{r}}
\toprule\toprule
Method
& Avg.
& ARC-C
& ARC-E
& BBH
& BoolQ
& COPA
& HellaS.
& LAMB.
& MMLU
& OBQA
& PIQA
& RACE
& SciQ
& SST-2
& TQA
& Wino. \\
\midrule

\refcell{OLMoE baseline}
& \refcell{51.15}
& \refcell{40.44}
& \refcell{\textbf{69.65}}
& \refcell{6.36}
& \refcell{62.02}
& \refcell{78.00}
& \refcell{\textbf{64.84}}
& \refcell{50.05}
& \refcell{25.76}
& \refcell{41.00}
& \refcell{75.41}
& \refcell{35.02}
& \refcell{86.20}
& \refcell{52.64}
& \refcell{\textbf{21.86}}
& \refcell{58.01} \\

OLMoE-Loop baseline
& 50.08 & 39.08 & \textbf{69.02} & 6.34
& 56.79 & 75.00 & 62.55 & 48.17
& 24.61 & \textbf{43.60} & 75.79 & 35.31
& 86.10 & 51.72 & \textbf{20.89} & 56.20 \\

\midrule

OLMoE-Loop DuoMI
& \textbf{51.82} & \textbf{42.06} & 68.52 & \textbf{14.10}
& 61.10 & 74.00 & \textbf{63.04} & 49.54
& 24.60 & 42.80 & \textbf{75.84} & 34.26
& \textbf{86.70} & \textbf{61.47} & 20.58 & \textbf{58.64} \\

\midrule

ModuleFormer-style MI
& 50.01 & 39.33 & 67.47 & 7.66
& 62.02 & 71.00 & 60.95 & 48.24
& \textbf{26.58} & 40.40 & 74.43 & 34.93
& 86.20 & 53.10 & 19.97 & 57.93 \\

Path-style shared router
& 50.76 & 39.59 & 68.64 & 7.39
& 59.66 & \textbf{79.00} & 62.01 & 50.24
& 24.60 & 38.20 & 74.43 & 34.93
& \textbf{86.70} & 58.72 & 20.01 & 57.22 \\

RMoE-style recurrent router
& 50.75 & 39.16 & 68.27 & 12.10
& \textbf{63.18} & \textbf{79.00} & 61.67 & \textbf{50.40}
& 25.99 & 41.00 & 74.97 & \textbf{35.79}
& 83.80 & 49.31 & 20.08 & 56.59 \\

\bottomrule\bottomrule
\end{tabular}%
}

\endgroup
\end{table*}

As shown in Table~\ref{tab:duomi_main}, DuoMI raises the average score
of the OLMoE-Loop baseline from $50.08$ to $51.82$, a gain of $1.74$
percentage points. For reference, OLMoE without cross-layer expert
sharing scores $51.15$. Thus, the looped baseline scores below the
unshared model in this comparison, whereas DuoMI exceeds it by $0.67$
points.

DuoMI improves on the looped baseline on $9$ of the $15$ tasks.
The largest gains occur on SST-2 (from $51.72$ to $61.47$) and BBH
(from $6.34$ to $14.10$). Even when both tasks are excluded, the mean
score across the remaining $13$ tasks increases by approximately $0.66$
points. The gains vary by task: DuoMI improves ARC-C and BoolQ by
$2.98$ and $4.31$ points, respectively, but scores $1.00$ point lower
on COPA and $1.05$ points lower on RACE.

Among the three adapted routing methods, the Path-style shared router
achieves the highest average score ($50.76$), closely followed by the
RMoE-style recurrent router ($50.75$). ModuleFormer-style MI scores
$50.01$. DuoMI exceeds the strongest of these alternatives by $1.06$
points. On this backbone, directly regularizing how the two rounds
use a shared pool yields a higher average score than optimizing
input--expert mutual information within each round, sharing router
parameters across rounds, or passing a routing state between them.

\subsection{Ablation Studies}
\label{sec:ablation}

\begin{table*}[t]
\centering
\caption{
Ablation study of Dual Mutual Information Regularization.
Bold indicates the best score in each column among the looped models.
OLMoE baseline is shown in gray as an additional reference;
its scores are bolded only when they attain the overall column maximum.
}
\label{tab:duomi_ablation}

\begingroup
\setlength{\tabcolsep}{2pt}
\renewcommand{\arraystretch}{1.15}
\small

\resizebox{\textwidth}{!}{%
\begin{tabular}{l*{16}{r}}
\toprule\toprule
Method
& Avg.
& ARC-C
& ARC-E
& BBH
& BoolQ
& COPA
& HellaS.
& LAMB.
& MMLU
& OBQA
& PIQA
& RACE
& SciQ
& SST-2
& TQA
& Wino. \\
\midrule

\refcell{OLMoE baseline}
& \refcell{51.15}
& \refcell{40.44}
& \refcell{69.65}
& \refcell{6.36}
& \refcell{62.02}
& \refcell{\textbf{78.00}}
& \refcell{\textbf{64.84}}
& \refcell{50.05}
& \refcell{\textbf{25.76}}
& \refcell{41.00}
& \refcell{75.41}
& \refcell{35.02}
& \refcell{86.20}
& \refcell{52.64}
& \refcell{\textbf{21.86}}
& \refcell{58.01} \\

OLMoE-Loop baseline
& 50.08 & 39.08 & 69.02 & 6.34
& 56.79 & 75.00 & 62.55 & 48.17
& 24.61 & \textbf{43.60} & 75.79 & \textbf{35.31}
& 86.10 & 51.72 & \textbf{20.89} & 56.20 \\

\midrule

OLMoE-Loop DuoMI
& \textbf{51.82} & \textbf{42.06} & 68.52 & \textbf{14.10}
& 61.10 & 74.00 & \textbf{63.04} & 49.54
& 24.60 & 42.80 & \textbf{75.84} & 34.26
& 86.70 & 61.47 & 20.58 & 58.64 \\

\midrule

$G$-only
& 50.53 & 41.98 & \textbf{70.62} & 5.76
& 61.10 & 73.00 & 62.78 & 49.33
& 24.85 & 41.60 & 75.19 & 34.07
& 86.30 & 50.80 & 19.87 & \textbf{60.77} \\

$D$-band only
& 51.46 & 40.87 & 69.02 & 4.91
& 61.77 & 76.00 & 62.12 & 49.35
& 24.78 & 40.60 & 75.24 & 33.78
& \textbf{87.40} & \textbf{68.12} & 19.86 & 58.01 \\

w/o $D$ lower bound
& 51.05 & 39.68 & 67.85 & 10.63
& \textbf{62.45} & \textbf{78.00} & 62.88 & \textbf{50.48}
& \textbf{25.35} & 43.40 & 75.30 & 35.12
& 86.50 & 50.69 & 20.46 & 56.91 \\

w/o $D$ upper bound
& 51.35 & 38.91 & 69.49 & 13.50
& 59.88 & 76.00 & 63.01 & 49.72
& 25.09 & 40.20 & 75.14 & 33.97
& 84.70 & 63.30 & 20.39 & 56.91 \\

\bottomrule\bottomrule
\end{tabular}%
}

\endgroup
\end{table*}

\paragraph{Round-level and token-level constraints.}
Table~\ref{tab:duomi_ablation} compares the contributions of the two
constraints. Starting from the looped baseline score of $50.08$,
constraining only the difference in aggregate expert usage between
rounds ($G$-only) raises the average to $50.53$. Constraining only
the degree of cross-round routing change for individual tokens
($D$-band only) raises it to $51.46$. The token-level constraint yields
the larger gain on its own, but adding the $G$ constraint improves the
average further, to $51.82$. Thus, even when token-level routing change
is already controlled, explicitly limiting the dependence of aggregate
expert usage on the round provides an additional benefit in this setting.

\paragraph{Does token-level routing change require a two-sided constraint?}
We retain the $G$ constraint and remove either the lower or the upper
bound on $\widetilde D$. Removing the lower bound leaves only a penalty
for excessive cross-round change and yields an average score of $51.05$.
Removing the upper bound leaves only a penalty for insufficient change
and yields $51.35$. These scores are $0.77$ and $0.47$ points below the
full model, respectively. In the settings tested, encouraging some
change in expert selection while limiting excessive replacement
therefore performs better on average than penalizing either extreme
alone.

\subsection{Cross-Round Routing Analysis}
\label{sec:routing_analysis}

Downstream scores do not reveal which routing decisions have changed.
We therefore track both the soft statistics used in the training
objective, $\widetilde G$ and $\widetilde D$, and expert retention
under actual Top-$k$ routing. We compute all three quantities on the
same groups and valid token positions. For group $b$ and shared expert
pool $j$, we define token-level expert retention as
\begin{equation}
O^{\mathrm{token}}_{b,j}
=
\frac{1}{N_b}
\sum_{i=1}^{N_b}
\frac{
    \left|S_{j,1}(x_i)\cap S_{j,2}(x_i)\right|
}{k}.
\label{eq:exp_token_retention}
\end{equation}
For example, under Top-$8$ routing, a token that retains two of its
eight experts across rounds has a retention rate of $25\%$. This metric
measures overlap between the experts actually selected and cannot be
recovered directly from the soft statistic $\widetilde D$.

\begin{figure*}[t]
    \centering
    \includegraphics[width=0.8\textwidth]{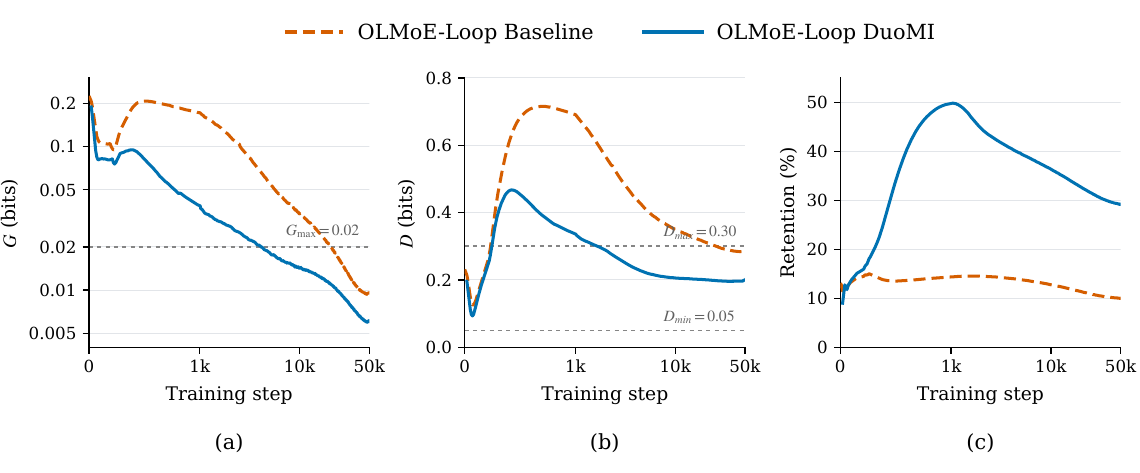}
    \caption{
        Cross-round routing trajectories for the OLMoE-Loop baseline
        and DuoMI under its default configuration. From left to right:
        round-level marginal mutual information $\widetilde G$,
        token-level conditional mutual information $\widetilde D$,
        and the fraction of Top-$8$ experts retained for the same
        token across rounds. Horizontal dashed lines mark the
        thresholds applied to the soft statistics during training.
    }
    \label{fig:duomi_routing_dynamics}
\end{figure*}

Figure~\ref{fig:duomi_routing_dynamics}(a) shows that $\widetilde G$
falls below $0.02$ for both models by the end of training, but their
trajectories differ. DuoMI crosses the threshold midway through
training, while aggregate expert usage in the baseline remains more
strongly dependent on the round for longer. Comparing only the final
checkpoints would therefore understate the difference in routing
behavior during training.

At the token level, the difference is visible in both the soft routing
distributions and the selected expert sets.
In Figure~\ref{fig:duomi_routing_dynamics}(b), the baseline's
$\widetilde D$ peaks at approximately $0.70$, whereas DuoMI's peaks
at approximately $0.47$ and returns to the target band earlier.
In our configuration, the logged values of $\widetilde D$ indicate
that the lower-bound penalty was active mainly within the first
$100$ training steps. Nevertheless, removing this bound lowers the
average downstream score by $0.77$ points
(Table~\ref{tab:duomi_ablation}). We therefore retain it throughout
training: after the initial steps, it serves as a safeguard that incurs
no penalty unless token-level routing change falls below the lower
threshold.

Figure~\ref{fig:duomi_routing_dynamics}(c) shows that, by the end of
training, the expert retention rate is approximately $10\%$ for the
baseline and $29\%$ for DuoMI. Under Top-$8$ routing, this corresponds
to an increase from fewer than one retained expert per token, on average,
to approximately $2.3$. Thus, in the regime where the baseline replaces
nearly all of a token's experts, DuoMI retains more of them across
rounds, while most experts selected in the second round still differ
from those selected in the first.

\subsection{Sensitivity to Constraint Thresholds and Regularization Strength}
\label{sec:sensitivity}

\begin{table*}[ht]
\centering
\caption{
Sensitivity to regularization thresholds and weight.
The default DuoMI setting uses
$\tilde g_{\max}=0.02$,
$[\tilde d_{\min},\tilde d_{\max}]=[0.05,0.30]$,
and $\lambda_G=0.1$.
Intervals denote target bands for $\widetilde D$.
Bold highlights the top two entries in each column among the looped
models, including ties at the cutoff.
OLMoE baseline is shown in gray as an additional reference;
its scores are bolded only when they attain the overall column maximum.
}
\label{tab:duomi_thresholds}

\begingroup
\setlength{\tabcolsep}{2pt}
\renewcommand{\arraystretch}{1.15}
\small

\resizebox{\textwidth}{!}{%
\begin{tabular}{l*{16}{r}}
\toprule\toprule
Method / Setting
& Avg.
& ARC-C
& ARC-E
& BBH
& BoolQ
& COPA
& HellaS.
& LAMB.
& MMLU
& OBQA
& PIQA
& RACE
& SciQ
& SST-2
& TQA
& Wino. \\
\midrule

\refcell{OLMoE baseline}
& \refcell{51.15}
& \refcell{40.44}
& \refcell{\textbf{69.65}}
& \refcell{6.36}
& \refcell{62.02}
& \refcell{\textbf{78.00}}
& \refcell{\textbf{64.84}}
& \refcell{50.05}
& \refcell{25.76}
& \refcell{41.00}
& \refcell{75.41}
& \refcell{35.02}
& \refcell{86.20}
& \refcell{52.64}
& \refcell{\textbf{21.86}}
& \refcell{58.01} \\

OLMoE-Loop baseline
& 50.08 & 39.08 & \textbf{69.02} & 6.34
& 56.79 & 75.00 & 62.55 & 48.17
& 24.61 & \textbf{43.60} & 75.79 & \textbf{35.31}
& 86.10 & 51.72 & \textbf{20.89} & 56.20 \\

\midrule

OLMoE-Loop DuoMI
& \textbf{51.82} & \textbf{42.06} & 68.52 & \textbf{14.10}
& 61.10 & 74.00 & \textbf{63.04} & 49.54
& 24.60 & 42.80 & \textbf{75.84} & 34.26
& \textbf{86.70} & 61.47 & \textbf{20.58} & \textbf{58.64} \\

\shortstack[l]{
$\tilde g_{\max}=0.02$\\
$\widetilde D\in[0.20,0.30]$
}
& 51.15 & 40.27 & \textbf{69.19} & \textbf{13.42}
& 62.20 & 73.00 & 62.46 & 49.97
& \textbf{25.96} & 40.00 & 74.92 & 32.82
& 86.20 & 59.86 & 19.06 & 57.93 \\

\shortstack[l]{
$\tilde g_{\max}=0.02$\\
$\widetilde D\in[0.05,0.35]$
}
& \textbf{52.03} & \textbf{40.61} & 68.73 & 9.89
& 60.76 & 77.00 & 62.64 & 49.78
& 24.85 & 42.80 & 75.35 & 33.97
& 86.00 & \textbf{70.53} & 19.84 & 57.70 \\

\shortstack[l]{
$\tilde g_{\max}=0.02$\\
$\widetilde D\in[0.05,0.325]$
}
& 51.06 & 39.93 & 67.38 & 5.01
& 61.04 & 71.00 & \textbf{62.91} & \textbf{50.30}
& 25.66 & 42.60 & \textbf{76.39} & 34.26
& 85.50 & 66.54 & 18.60 & \textbf{58.80} \\

\shortstack[l]{
$\tilde g_{\max}=0.02$\\
$\widetilde D\in[0.05,0.40]$
}
& 51.51 & 39.68 & 67.00 & 7.65
& \textbf{62.94} & \textbf{78.00} & 62.55 & \textbf{50.46}
& 25.61 & \textbf{43.00} & 74.37 & 34.64
& 85.90 & \textbf{63.65} & 20.46 & 56.67 \\

\shortstack[l]{
$\tilde g_{\max}=0.015$\\
$\widetilde D\in[0.05,0.30]$
}
& 51.23 & 40.44 & 67.85 & 12.81
& \textbf{62.48} & 73.00 & 62.80 & 49.74
& 25.32 & 40.20 & 75.19 & \textbf{35.02}
& \textbf{88.70} & 59.06 & 18.95 & 56.91 \\

\shortstack[l]{
$\tilde g_{\max}=0.02,\;\lambda_G=1$\\
$\widetilde D\in[0.05,0.30]$
}
& 51.21 & 39.33 & 67.68 & 11.75
& 61.77 & \textbf{78.00} & 62.51 & 49.80
& \textbf{26.81} & 41.60 & 75.08 & 34.74
& 84.70 & 55.28 & 20.47 & \textbf{58.64} \\

\bottomrule\bottomrule
\end{tabular}%
}

\endgroup
\end{table*}

\paragraph{Target band for token-level routing change.}
Starting from the default band $[0.05,0.30]$,
Table~\ref{tab:duomi_thresholds} varies one threshold or regularization
coefficient at a time. With the upper bound fixed at $0.30$, raising
the lower bound from $0.05$ to $0.20$ reduces the average score from
$51.82$ to $51.15$. With the lower bound fixed at $0.05$, upper bounds
of $0.30$, $0.325$, $0.35$, and $0.40$ yield average scores of $51.82$,
$51.06$, $52.03$, and $51.51$, respectively. Performance therefore
does not vary monotonically as the upper bound is relaxed.

The band $[0.05,0.35]$ yields the highest average in this sweep,
exceeding the default by $0.21$ points. This gain reflects a trade-off
across tasks: in a direct comparison, the default band scores higher
on nine tasks, the wider band scores higher on five, and the two tie
on one. We therefore report $0.35$ as the best-performing upper bound
by average score in this sweep, rather than as a universally optimal
choice.

\paragraph{Threshold and strength of the round-level constraint.}
With the default target band for $\widetilde D$, tightening
$\tilde g_{\max}$ from $0.02$ to $0.015$ reduces the average score
from $51.82$ to $51.23$. Keeping the thresholds fixed and increasing
$\lambda_G$ from $0.1$ to $1$ yields an average of $51.21$.
Together with the $G$-only and $D$-band only ablations, these results
indicate that including the round-level constraint is beneficial in
this configuration, but neither tightening its threshold nor increasing
its penalty weight yields a further gain. The corresponding cross-round
routing trajectories, including actual Top-$8$ expert retention under
different threshold settings, are shown in
Appendix Figure~\ref{fig:duomi_threshold_dynamics}.

\section{Conclusion}
\label{sec:conclusion}

We propose Dual Mutual Information Regularization (DuoMI), which uses round-level marginal mutual information to limit the dependence of aggregate expert usage on the invocation round, and token-level conditional mutual information to encourage the average degree of cross-round routing change to remain within a prescribed interval. It also uses conditional mutual information given the token to keep the average change in routing within a target range. Both terms are computed from router outputs during training, add no model parameters, and leave sparse top-$k$ routing unchanged at inference.
On a looped OLMoE backbone that shares experts between paired adjacent layers, DuoMI raised the average score across $15$ downstream tasks from $50.08\%$ to $51.82\%$, a gain of $1.74$ percentage points over the looped baseline. The full method outperformed variants that used either constraint alone, and removing either bound on token-level routing change also reduced the average score. These results suggest that, in the architecture we studied, encouraging similar expert usage across rounds at the aggregate level is compatible with allowing individual tokens to change which experts they use.
Our compute budget limited evaluation to a looped backbone derived from OLMoE-1B-7B. Whether the gains generalize to other open-source architectures or larger models remains to be tested. Promising future directions include: (i) formulating mutual information constraints on expert routing when a shared pool is used across more than two rounds, and (ii) developing new looped Transformer architectures whose rounds differ by design, so that the architecture can better support the behaviors targeted by new training objectives.

\clearpage
\makeatletter
\global\@colroom\@colht
\global\vsize\@colht
\makeatother




\bibliography{example_paper}
\bibliographystyle{icml2027}

\newpage
\appendix
\onecolumn

\section{Theoretical Properties of Dual Mutual Information}
\label{app:theory}

Throughout this section, we fix a statistical group $b$ containing
$N_b>0$ valid token positions and a shared expert pool $j$.
All logarithms are taken to base $2$.

\subsection{Conditional Mutual Information and Expert Replacement}
\label{app:hard_replacement}

We prove that, under the uniform distributions over hard Top-$k$
selections in Eq.~\eqref{eq:duomi_hard_distribution}, the conditional
mutual information $D_{b,j}$ is exactly the mean fraction of experts
replaced between rounds.

Fix a token-context pair $x_i$. For brevity, write
$S_r=S_{j,r}(x_i)$ and $q_r=q_{j,r}(\cdot\mid x_i)$ for $r\in\{1,2\}$.
Both sets have size $k$. Let $m_i=|S_1\cap S_2|$ and
$u=(q_1+q_2)/2$. For an expert selected in both rounds,
$q_1(e)=q_2(e)=u(e)=1/k$, so its contribution to either KL divergence
is zero. For an expert selected only in round~$1$,
$q_1(e)=1/k$ and $u(e)=1/(2k)$. There are $k-m_i$ such experts, giving
\begin{equation}
\operatorname{KL}_2(q_1\Vert u)
=
\frac{k-m_i}{k}
\log_2\!\left(\frac{1/k}{1/(2k)}\right)
=
1-\frac{m_i}{k}.
\label{eq:app_kl_replacement}
\end{equation}
The same calculation for round~$2$ gives
$\operatorname{KL}_2(q_2\Vert u)=1-m_i/k$. Their average is therefore
$\operatorname{JSD}_2(q_1,q_2)=1-m_i/k$. Averaging over the valid
token positions yields
\begin{equation}
D_{b,j}
=
\frac{1}{N_b}\sum_{i=1}^{N_b}
\left(1-\frac{|S_{j,1}(x_i)\cap S_{j,2}(x_i)|}{k}\right)
=
1-O^{\mathrm{token}}_{b,j},
\label{eq:app_replacement_retention}
\end{equation}
where $O^{\mathrm{token}}_{b,j}$ is the hard expert retention rate
defined in Eq.~\eqref{eq:exp_token_retention}.

This identity relies on assigning equal mass to each of the $k$
selected experts. The soft routing distributions used during training
instead assign probabilities according to the router logits over all
$K$ experts. Consequently, $1-\widetilde D_{b,j}$ cannot be interpreted
as the fraction of Top-$k$ experts retained.

\subsection{Relation Between Marginal and Conditional Mutual Information}
\label{app:mi_relation}

Under the empirical distribution in Eq.~\eqref{eq:duomi_joint_distribution},
the round $R$ is sampled uniformly from $\{1,2\}$ independently of the
token position $X$. Expanding $I(R;X,E)$ in two orders by the chain rule
for mutual information~\citep{cover2006elements} gives
\begin{equation}
I(R;X)+I(R;E\mid X)
=
I(R;E)+I(R;X\mid E).
\label{eq:app_chain_rule}
\end{equation}
Since $I(R;X)=0$, substituting
$G_{b,j}=I(R;E)$ and $D_{b,j}=I(R;E\mid X)$ yields
\begin{equation}
D_{b,j}-G_{b,j}
=
I(R;X\mid E)
\geq 0.
\label{eq:app_information_gap}
\end{equation}
Moreover,
$D_{b,j}=H(R\mid X)-H(R\mid X,E)\leq H(R\mid X)=1$ bit.
Together with the nonnegativity of $G_{b,j}$, this proves
\begin{equation}
0\leq G_{b,j}\leq D_{b,j}\leq 1.
\label{eq:app_information_bounds}
\end{equation}

The gap in Eq.~\eqref{eq:app_information_gap} has a direct interpretation:
it measures how much additional information the token position provides
about the round once the expert identity is known. For example, an
expert may be selected equally often in both rounds but serve different
tokens in each. Such changes can cancel in the aggregate expert counts
while remaining visible in the token-level comparison.

The same proof applies when $E$ is sampled from the soft distribution
$p_{j,r}(\cdot\mid X)$: replacing $q_{j,r}$ with $p_{j,r}$ changes
the conditional distribution of $E$, but leaves $R$ uniform and
independent of $X$. Hence the soft statistics used for training also
satisfy $0\leq\widetilde G_{b,j}\leq\widetilde D_{b,j}\leq1$.

\section{Experimental and Implementation Details}
\label{app:experimental_details}

\subsection{Model Architecture, Pretraining, and Evaluation}
\label{app:training_eval}

\paragraph{Architecture.}
Our looped backbone is derived from OLMoE-1B-7B~\citep{muennighoff2025olmoe}.
Numbering the $16$ Transformer layers from $1$ to $16$, layers
$1$, $2$, $15$, and $16$ each have their own expert pool.
The middle $12$ layers form six pairs:
\[
(3,4),\quad (5,6),\quad (7,8),\quad
(9,10),\quad (11,12),\quad (13,14).
\]
Within each pair, the earlier layer is round~$1$ and the later layer
is round~$2$. The two layers share the parameters of all $64$ expert
FFNs, while their attention, router, and normalization parameters
remain independent. Each invocation selects $8$ experts per token.
Thus, the $16$ layers contain $10$ distinct expert pools, of which
$M=6$ are invoked twice and enter the DuoMI objective. Counting
shared weights only once, the looped backbone has approximately
$4.5$ billion trainable parameters, with approximately $1.3$ billion
parameters active per token.

\paragraph{Pretraining.}
We train the looped models from scratch in MindSpeed-LLM using the
official \texttt{sample-100BT} subset of FineWeb-Edu
v1.4.0~\citep{penedo2024fineweb}. Each optimizer update processes a
global batch of $512$ sequences of length $4{,}096$. Over $50{,}000$
updates, this corresponds to
\[
50{,}000 \times 512 \times 4{,}096
=104{,}857{,}600{,}000
\]
training token positions. This is the number of positions processed,
before excluding masked positions; it is not a count of unique corpus
tokens.

\paragraph{Downstream evaluation.}
All models use the same evaluation protocol. The $15$ benchmarks in
Table~\ref{tab:duomi_main} use three scoring metrics:
\begin{itemize}
    \item \textbf{Normalized accuracy:}
    ARC-Challenge, ARC-Easy, HellaSwag, OpenBookQA, PIQA, and SciQ.
    \item \textbf{Accuracy:}
    BoolQ, COPA, LAMBADA (\texttt{lambada\_openai}), MMLU,
    RACE, SST-2, and WinoGrande.
    \item \textbf{Exact match:}
    BBH and TriviaQA.
\end{itemize}
We express each task score as a percentage and compute the reported
average as the unweighted arithmetic mean of the $15$ scores. Each
task therefore has equal weight regardless of its evaluation set size.

\subsection{Computation of the DuoMI Objective}
\label{app:duomi_implementation}

\paragraph{Statistical groups and token alignment.}
Each mutual information estimate uses a fixed group of eight training
sequences. Groups are formed within a single microstep, i.e., one
forward and backward pass before the optimizer update. For example,
with a local microbatch of two sequences, four data-parallel ranks
contribute two sequences each to the same group. At sequence length
$4{,}096$, a group contains at most $32{,}768$ valid positions.
We exclude padding and positions masked out of the language-modeling
loss and use exactly the same valid positions in both rounds.
Alignment is by sequence and token position, so it does not require
the hidden states at the two layers to be identical.

\paragraph{Computing the two statistics.}
We obtain the router logits before expert-parallel dispatch, so
dispatching a token to multiple selected experts does not cause that
token to be counted multiple times. For a fixed group $b$ and pool
$j$, let
$p_{i,r}=\operatorname{softmax}(z_{j,r}(x_i)/\tau)$ denote the
auxiliary distribution over all $64$ experts. The quantities needed
from each rank are the number of valid positions, the sum of the
routing distributions in each round, and the sum of the per-token
divergences. After summing these contributions over the ranks in the
group, write
\begin{equation}
\mathbf{s}_r=\sum_{i=1}^{N_b}p_{i,r},
\qquad
u=\sum_{i=1}^{N_b}\operatorname{JSD}_2(p_{i,1},p_{i,2}).
\end{equation}
The two group statistics are then
\begin{equation}
\widetilde G_{b,j}
=
\operatorname{JSD}_2\!\left(
\frac{\mathbf{s}_1}{N_b},\frac{\mathbf{s}_2}{N_b}
\right),
\qquad
\widetilde D_{b,j}=\frac{u}{N_b}.
\label{eq:app_group_statistics}
\end{equation}
In particular, computing a separate marginal JSD on each rank and
then averaging those values would not, in general, recover
$\widetilde G_{b,j}$.

\paragraph{Penalty and reduction.}
We use $\tau=1$. The default marginal threshold is
$\widetilde g_{\max}=0.02$ bits, the conditional target band is
$[\widetilde d_{\min},\widetilde d_{\max}]=[0.05,0.30]$ bits,
and the three squared hinge penalties have weights
$\lambda_G=\lambda_L=\lambda_U=0.1$.
We evaluate the penalty $\Phi_{b,j}$ in Eq.~\eqref{eq:duomi_penalty}
separately for each group and shared pool. A global batch of $512$
sequences contains $512/8=64$ statistical groups. The contribution
of DuoMI to one optimizer update is therefore
\begin{equation}
\mathcal L_{\mathrm{DuoMI},t}
=
\frac{w_t}{64\times 6}
\sum_{b\in\mathcal G_t}\sum_{j=1}^{6}\Phi_{b,j}.
\label{eq:app_update_reduction}
\end{equation}
The order of operations matters: a threshold violation in one group
or pool incurs a penalty even if averaging the statistics across the
entire update would place them inside the target region. The auxiliary
distributions are used only for this training objective. Expert
dispatch and output weighting continue to use the original sparse
Top-$8$ router.

\section{Routing Dynamics Across Constraint Settings}
\label{app:routing_sensitivity}

\begin{figure}[t]
    \centering
    \includegraphics[
        width=0.96\linewidth,
        height=0.72\textheight,
        keepaspectratio
    ]{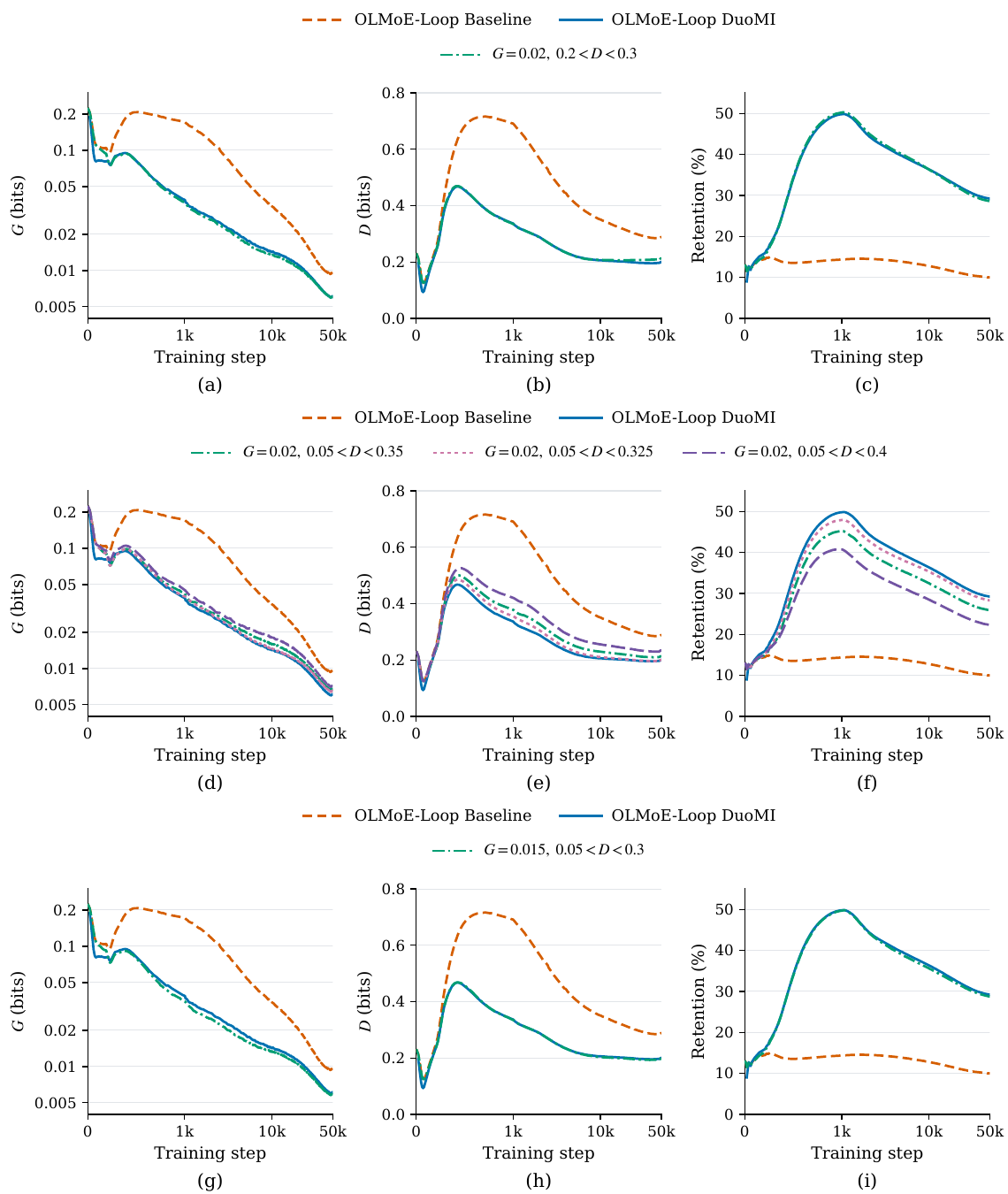}
    \caption{
        Cross-round routing under different constraint thresholds.
        Columns show, from left to right, round-level marginal mutual
        information $\widetilde G$ (bits), token-level conditional
        mutual information $\widetilde D$ (bits), and hard Top-$8$
        expert retention (\%). The orange dashed and blue solid curves
        denote the OLMoE-Loop baseline and default DuoMI configuration,
        respectively. Panels (a)--(c) raise $\widetilde d_{\min}$
        from $0.05$ to $0.20$; panels (d)--(f) compare
        $\widetilde d_{\max}\in\{0.30,0.325,0.35,0.40\}$; and panels
        (g)--(i) lower $\widetilde g_{\max}$ from $0.02$ to $0.015$.
        All other thresholds and penalty weights remain at their
        default values.
    }
    \label{fig:duomi_threshold_dynamics}
\end{figure}

Figure~\ref{fig:duomi_threshold_dynamics} complements the downstream
scores in Table~\ref{tab:duomi_thresholds} with routing trajectories
from the same threshold sweep. Each row varies one threshold and
includes both the unregularized OLMoE-Loop baseline and the default
DuoMI run as references. The first two columns show the soft
statistics used in the training objective. The third shows the
fraction of the eight experts selected for a token in round~$1$
that are selected again for that token in round~$2$.

Raising the lower bound on $\widetilde D$ from $0.05$ to $0.20$
produces trajectories close to the default run in panels (a)--(c).
The two runs differ slightly in $\widetilde D$ near the end of
training, but their hard retention curves nearly coincide. These
aggregate curves do not establish how often the lower-bound penalty
was active: the penalty is evaluated separately for each group and
pool before averaging.

Varying the upper bound on $\widetilde D$ produces a clearer change
in actual expert selection. In panels (e) and (f), a higher upper
bound generally allows greater token-level divergence and yields
lower retention after the early training phase. At the end of
training, retention is approximately $29\%$ with the default upper
bound of $0.30$, compared with approximately $22\%$ with an upper
bound of $0.40$. Downstream scores do not follow the same ordering:
the intermediate upper bound of $0.35$ gives the highest average
score in this sweep (Table~\ref{tab:duomi_thresholds}). Thus,
increasing hard expert retention is not itself the downstream
objective.

Finally, tightening $\widetilde g_{\max}$ from $0.02$ to $0.015$
slightly lowers $\widetilde G$ over part of training in panel (g),
while $\widetilde D$ and hard retention remain close to the default
trajectories in panels (h) and (i). In this comparison, the clearest
change is in aggregate expert usage across rounds, with little change
in how many experts an individual token retains.

\end{document}